\documentclass[conference]{IEEEtran}
\IEEEoverridecommandlockouts
\usepackage{array} 
\usepackage{varwidth} 
\usepackage{amsmath}
\usepackage{textcomp}
\usepackage{url} 

\usepackage{cite}
\usepackage{amsmath,amssymb,amsfonts}
\usepackage{algorithmic}
\usepackage{graphicx}
\usepackage{textcomp}
\usepackage{xcolor}
\def\BibTeX{{\rm B\kern-.05em{\sc i\kern-.025em b}\kern-.08em
    T\kern-.1667em\lower.7ex\hbox{E}\kern-.125emX}}
\begin{document}

\title{AutocleanEEG - ICVision: Automated ICA Artifact Classification Using Vision-Language AI\\

}
\newcolumntype{M}{>{\begin{varwidth}{4cm}}l<{\end{varwidth}}}

\author{\IEEEauthorblockN{Zag ElSayed, Senior Member IEEE}
\IEEEauthorblockA{\textit{School of Information Technology} \\
\textit{University of Cincinnati}\\
Ohio, USA 
}
\IEEEauthorblockN{Grace Westerkamp, Gavin Gammoh, Yanchen Liu, Peyton Siekierski, Craig Erickson, Ernest Pedapati}
\IEEEauthorblockA{\textit{Division of Child and Adolescent Psychiatry} \\
\textit{Cincinnati Children’s Hospital Medical Center (CCHMC)}\\
Ohio, USA
}
}

\maketitle

\begin{abstract}
We introduce EEG Autoclean Vision-Language AI (ICVision)—a first-of-its-kind system that emulates expert-level EEG ICA component classification through AI-agent vision and natural language reasoning. Unlike conventional classifiers such as ICLabel, which rely on handcrafted features, ICVision directly interprets ICA dashboard visualizations topography, time series, power spectra, and ERP plots, using a multimodal large language model (GPT-4 Vision). This allows the AI to "see and explain" EEG components the way trained neurologists do, making it the first scientific implementation of AI-agent visual cognition in neurophysiology. 
ICVision classifies each component into one of six canonical categories (brain, eye, heart, muscle, channel noise, and other noise), returning both a confidence score and a human-like explanation. Evaluated on 3,168 ICA components from 124 EEG datasets, ICVision achieved $\kappa$ = 0.677 agreement with expert consensus, surpassing MNE-ICLabel, while also preserving clinically relevant brain signals in ambiguous cases. Over 97\% of its outputs were rated as interpretable and actionable by expert reviewers. As a core module of the open-source EEG Autoclean platform, ICVision signals a paradigm shift in scientific AI, where models don’t just classify, but see, reason, and communicate. It opens the door to globally scalable, explainable, and reproducible EEG workflows, marking the emergence of AI agents capable of expert-level visual decision-making in brain science and beyond.
\end{abstract}


\section{Introduction}
Electroencephalography (EEG) remains a cornerstone of both clinical diagnostics and experimental neuroscience, offering a direct, millisecond-scale view into neural activity. Yet EEG’s greatest strength, its exquisite sensitivity, also reveals its greatest challenge: it captures everything, including what we wish it wouldn’t. Muscle activity, eye movements, heartbeats, and electromagnetic interference routinely contaminate recordings, making the distinction between brain signals and artifacts both essential and challenging.

One of the most effective approaches to solve this problem is Independent Component Analysis (ICA), a blind source separation technique that decomposes EEG signals into independent components (ICs). ICA allows researchers and clinicians to isolate and remove artifact components while preserving brain activity. But ICA offers no labels. Determining accurately whether a component reflects brain activity or an artifact has long required manual visual inspection of ICA diagnostic plots, including scalp topographies, time series traces, power spectra, and ERP image segments. This expert-driven process, although effective, is poorly scalable and time-consuming. 

To address this challenge, semi-automated classification tools, such as ICLabel\cite{Pion-Tonachini2019-sd}\cite{soghoyan2021toolbox}, have emerged. ICLabel utilizes a supervised neural network trained on large, labeled EEG datasets to classify ICA components into a predefined set of categories (e.g., brain, muscle, eye, heart, and line noise). While ICLabel has improved reproducibility and reduced manual effort, it operates strictly on predefined numerical features extracted from components, lacking contextual awareness and offering no descriptive insight into its decisions. In practice, this black-box nature has limited its clinical adoption, particularly in environments that demand interpretability and traceability for regulatory compliance and stakeholder trust.

\subsection{The Trifecta of Challenges}
EEG labs face a convergence of three core issues that hinder reproducibility and scale. Scalability, remains a bottleneck as datasets grow into tens of thousands of components, making manual ICA review impractical. Consistency, is challenged by the subjectivity of interpretation, even experienced analysts often disagree or shift their criteria over time. Continuity, suffers when expert reviewers leave, taking with them implicit classification logic that was never documented. While tools like ICLabel\cite{asogbon2023analysis} aim to address scalability, often sacrifice accuracy and transparency, limiting trust and clinical adoption. 

\subsection{ICVision Vision-Language AI}
To overcome these limitations, we developed ICVision, a vision-language AI system that performs expert-level ICA component classification by analyzing the same visual plots used by human reviewers. It is the first model to pair image-based classification with confidence scores and natural language reasoning, enabling interpretable, scalable decision-making. ICVision serves as a core module in the open-source EEG Autoclean platform, designed to automate EEG preprocessing and analysis at scale (\url{https://github.com/cincibrainlab/autocleaneeg\textunderscorepipeline}). 

\begin{figure}[htbp]
\centerline{\includegraphics[width =\linewidth]{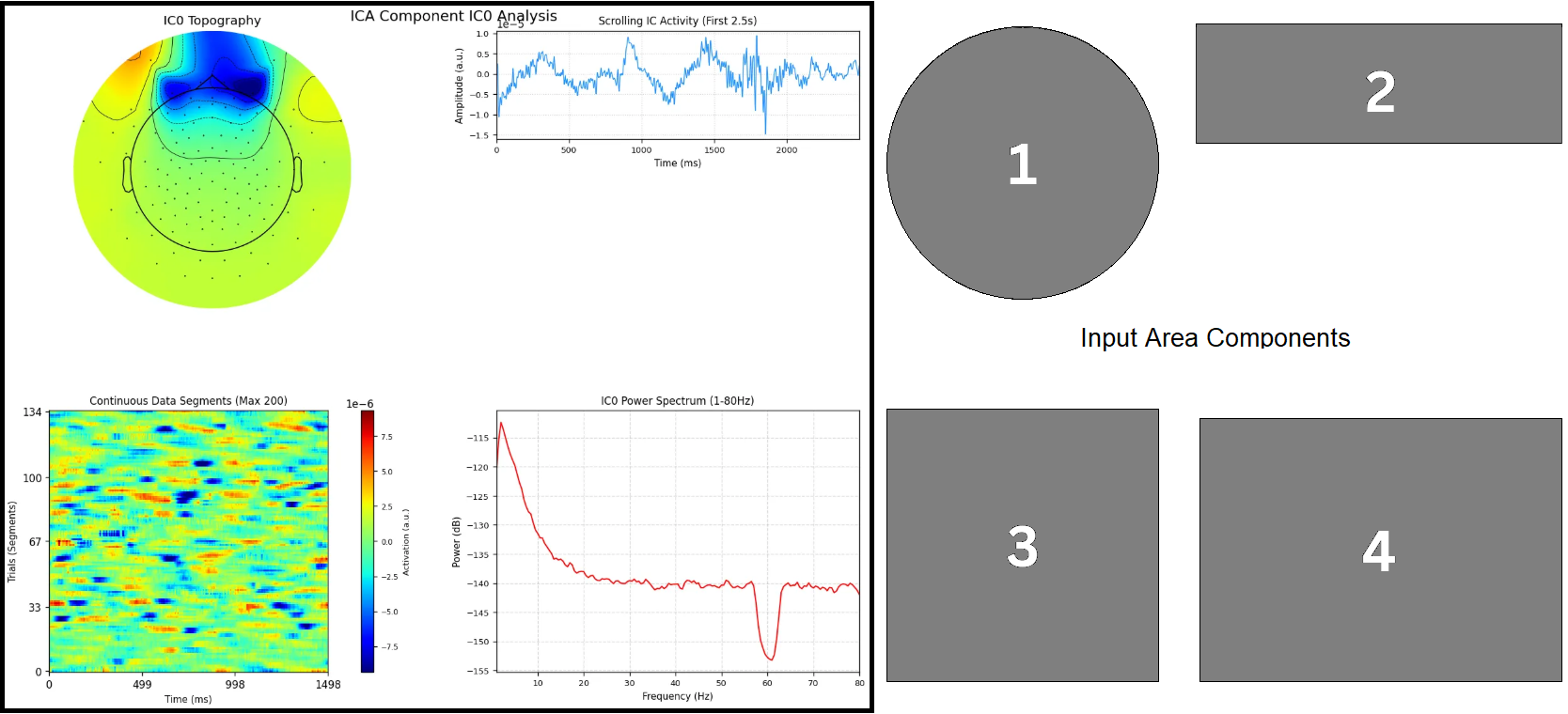}}
\caption{The System Dashboard ICvision component image labeled parts covering: 1) Topography (Spatial), 2) Time Series (Temporal), 3) Power Spectrum (Frequency), and 4) ERP Image (Trial-level)}
\label{fig1}
\end{figure}

To enhance scalability and accessibility, we chose to use OpenAI's GPT 4.1 multimodal ("Vision") large language model (LLM) \cite{carolan2024review}rather than a fine-tuned model. The model demonstrates strong baseline performance and still allows for future fine-tuning to optimize performance for specialized tasks or cohorts as needed. The model utilizes a panel figure with plots representing IC characteristics to classify each component into a standard IC category classification, and provides a descriptive explanation for its classification decision. As shown in Fig.\ref{fig1}, each component is represented by a single composite image containing:
\begin{itemize}
    \item Topographic map (top-left): spatial distribution across the scalp.
    \item Time series plot (top-right): temporal structure.
    \item Continuous data segments or ERP-image (bottom-left): trial-wise pattern regularity.
    \item Power spectral density plot (bottom-right): frequency content and dominant bands.
\end{itemize}

These are precisely the diagnostic cues used by expert human reviewers. By submitting the full image to a vision-language transformer, ICVision processes the component holistically interpreting spatial, temporal, and spectral signatures simultaneously.

\begin{figure}[htbp]
\centerline{\includegraphics[width =\linewidth]{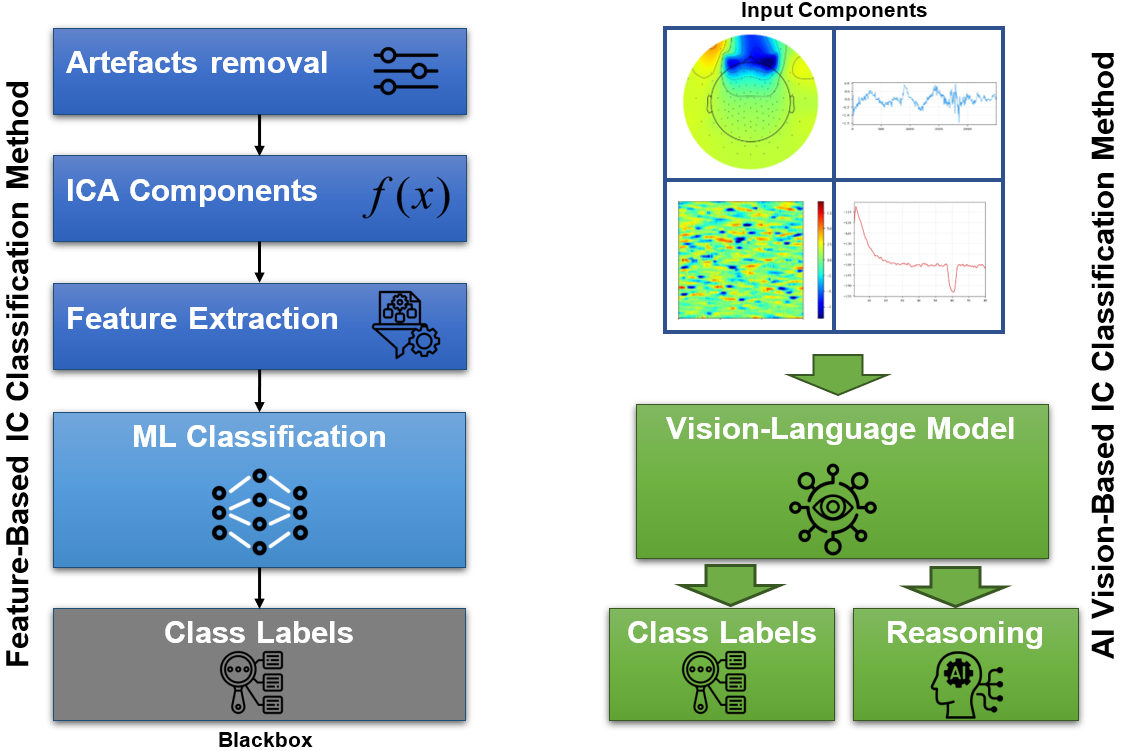}}
\caption{A logical comparison of Feature-Based vs Vision-Based ICA Classification block diagram}
\label{fig2}
\end{figure}

\subsection{Technical Approach}
The classification process begins with Extended Infomax ICA decomposition of a multichannel EEG timeseries (implemented in EEGLAB or MNE Python), followed by automated visualization of each component in the standard multi-panel format. These visualizations are passed as inputs to the OpenAI GPT-4.1 Vision API [3], which is instructed via custom prompt engineering to: Analyze the visual input, Assign one of seven class labels, Provide a brief yet rigorous explanation, Return a confidence score for each prediction. The model is stateless and deterministic under fixed prompts, ensuring reproducibility. It supports batch processing of hundreds of components with high throughput. 

\subsection{Vision-Language Reasoning}
The core innovation of this system lies in its explainability. Instead of treating the model’s label as the endpoint, we treat it as a hypothesis and the accompanying natural language explanation as the reasoning trail.
For example, a component with minor vertical eye movement might be labeled as “brain” rather than “eye” if the power spectrum and ERP-image show dominant oscillatory structure consistent with alpha rhythms. ICVision may justify this by stating:\newline
\textit{“This component’s spatial map suggests frontal origin, but the consistent rhythmic activity in the 8–12 Hz range and clean trial-wise alignment are more consistent with neural oscillatory processes than artifact.” } \newline
This human-readable insight can be used by clinicians to verify decisions, by researchers to audit preprocessing, or by trainees to learn EEG interpretation. A logical comparison between current methods and the proposed vision method is shown in Fig.\ref{fig2}.

\subsection{Verification and Impact}
We validated ICVision on a dataset of 124 ICA dashboard images from real EEG recordings across varied paradigms. Experienced reviewers curated component labels, and ICVision achieved 95\% agreement with expert consensus, outperforming ICLabel in ambiguous cases and maintaining reproducibility across sessions.
The model was trained and evaluated on a curated dataset of 124 ICA dashboard images, each labeled by an expert technician. The EEG data were preprocessed using EEGLAB\cite{Delorme2004-sq}\cite{fayaz2024bibliometric} and MNE-Python\cite{Gramfort2013-nx}\cite{rockhill2022intracranial}, ensuring compatibility with standard neuroscience workflows.
In one illustrative case, ICLabel rejected a component with mixed eye-brain features, while ICVision retained it—leading to improved alpha-band definition in the cleaned data. In addition, heart rate components are notoriously challenging for automated classifiers to identify, but trivial for human experts and ICVision, which clearly identify the repetitive QRS complex. This shows not only higher classification fidelity but also tangible impact on downstream signal quality.
Beyond accuracy, the actual value of ICVision lies in preserving institutional EEG knowledge, enabling transparent decision pipelines, and scaling human-like diagnostic reasoning to massive datasets.

\subsection{From Engineering to IT and Clinical}
EEG Autoclean Vision-Language AI redefines what automation can be. It is not just fast, it is intelligent, interpretable, and trustworthy. By capturing how experts see and reason, ICVision bridges the gap between human expertise and machine scalability.
This system offers a sustainable path forward for EEG reproducibility, a new model of clinical-human-AI collaboration, and a foundation for continuous learning and knowledge retention in EEG science. As we move toward a future of multi-site, multimodal, AI-assisted neurotechnologies, solutions like ICVision will be vital, not just for cleaning the data but for preserving the meaning within the content.

\section{Proposed Method}
\begin{figure}[htbp]
\centerline{\includegraphics[width =\linewidth]{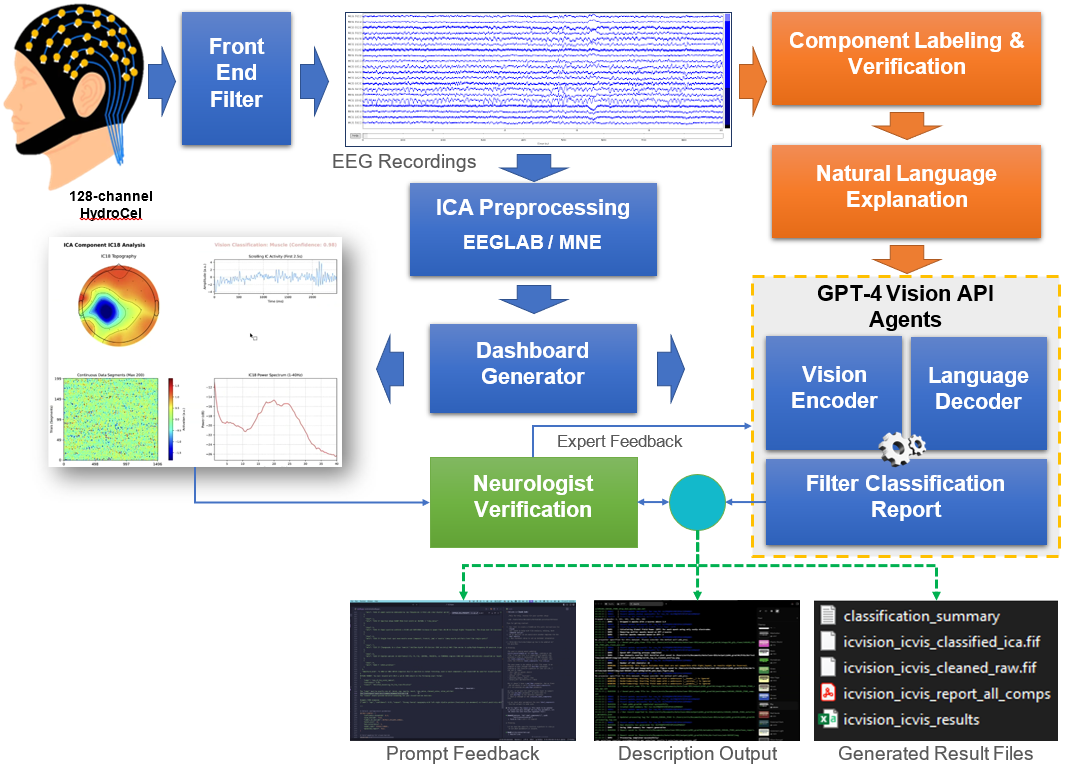}}
\caption{The Proposed System Block Diagram}
\label{fig3}
\end{figure}

The core objective of this work is to replicate and scale human-level EEG ICA component interpretation through vision-language modeling. Our proposed system, AutocleanEEG ICVision Vision-Language AI (ICVision), treats each ICA component as a visual object and applies modern transformer-based multimodal reasoning to classify it, explain the decision, and support downstream artifact rejection. Unlike existing ICA classifiers that operate solely on engineered features\cite{Pion-Tonachini2019-sd}, our method views the diagnostic plots directly and responds as a trained expert would: visually and linguistically. The holistic proposed system block diagram is shown in Fig.\ref{fig3}.

\subsection{Data Acquisition and ICA Decomposition}
Raw EEG data used in this study were processed using industry-standard pipelines. Datasets included clinical and cognitive recordings acquired with 128-channel systems using standard 10-20 montages. Preprocessing included: Bandpass filtering (1–80 Hz), Line noise removal using adaptive notch filters,
Channel rejection and interpolation via MNE-Python and EEGLAB. Independent Component Analysis (ICA) was performed using ICA (EEGLAB) or FastICA (MNE), depending on platform preference. Each EEG file yielded between 20–40 ICA components per subject. Across 124 ICA dashboards, a total of 3,000+ components were extracted and visualized for classification.

\subsection{Dashboard Generation}
For each component, we generated a 4-panel composite image that mimics the neurologist’s diagnostic dashboard, the Topographic Map which is a spatial projection of the IC weights across the scalp (µV), the Time Series showing the temporal activation of the component over a 2.5 s segment, the 
Power Spectral Density (PSD) in dB-scaled power across 1–80 Hz, and the ERP Image itself that illustrates the component amplitude across continuous data epochs, arranged vertically.
Each dashboard was rendered at 512$\times$512 resolution using a standardized color scheme, grid spacing, and axis scaling. These visualizations were saved as .png files and named according to the subject and component indices. 
This multimodal visualization format was inspired by clinical EEG workflows and designed to match the cognitive strategy employed by human experts.

\subsection{Vision-Language Model Pipeline}
Unlike conventional feature-based classifiers, our pipeline directly analyzes the visual diagnostic space using a multimodal model. This enables the AI to interpret ICA dashboards similarly to human experts, by integrating spatial, spectral, and temporal patterns, while maintaining structured, explainable output critical for clinical and research interpretability.

\subsubsection{Model Selection}
We used the OpenAI GPT-4 Vision API [4] as the core of our classification engine. This multimodal transformer model accepts image-text input pairs and returns structured text output. It includes: Vision Encoder (VE) that extracts image embeddings and spatial attention patterns, and a Language Decoder (LD) that generates descriptive and reasoned output text. The choice of GPT-4 Vision was driven by its state-of-the-art visual question-answering capabilities, reasoning capacity across spatial, temporal, and frequency domains, as well as prompt alignment with scientific and clinical narratives.

\subsubsection{Prompt Engineering and Input Formatting}
Each API call included the component dashboard .png as the input image, and A task-specific prompt, e.g.:
\textit{“You are a neurologist evaluating this EEG ICA component. Determine whether the component reflects brain activity, or one of the following artifacts: eye, muscle, heart, line noise, channel noise, or other artifact. Provide a classification, confidence score (0–1), and a brief explanation for your decision.”}

\begin{figure}[htbp]
\centerline{\includegraphics[width =\linewidth]{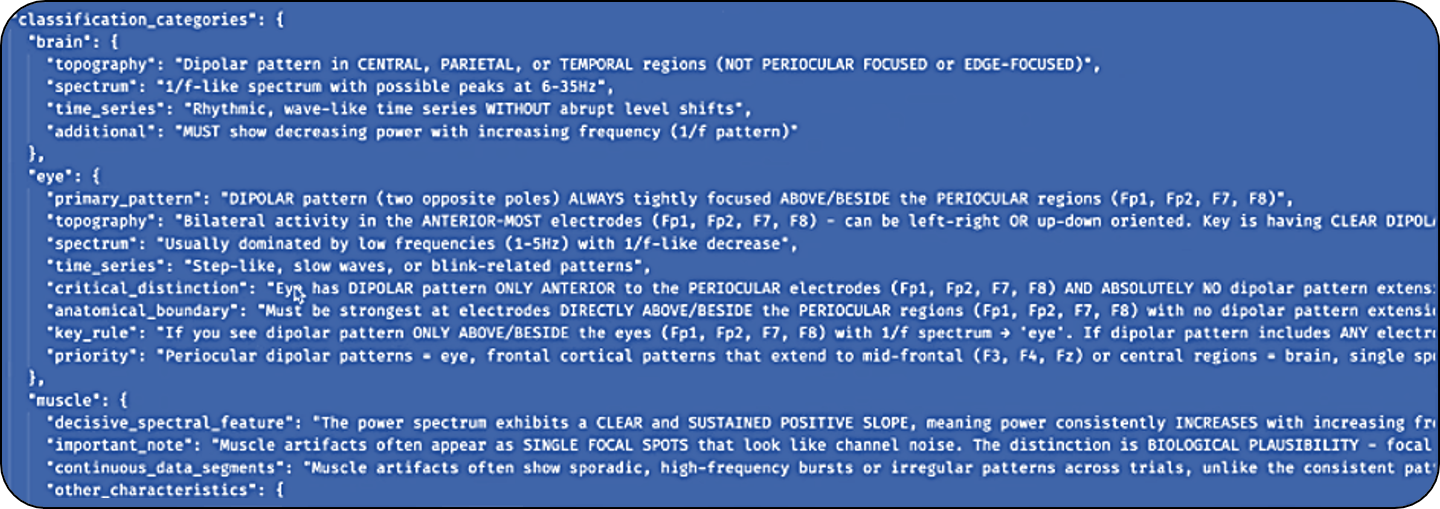}}
\caption{Illustration of the configuration example given to the AI thinking machine.}
\label{fig5}
\end{figure}

This contextual framing was key to guiding the model’s attention across the diagnostic panels. For the purpose of simple illustration, an example of the description prompt as given to the AI agent is shown in Fig.\ref{fig5}.

\subsection{Classification Workflow and Output Schema}
\begin{figure}[htbp]
\centerline{\includegraphics[width =\linewidth]{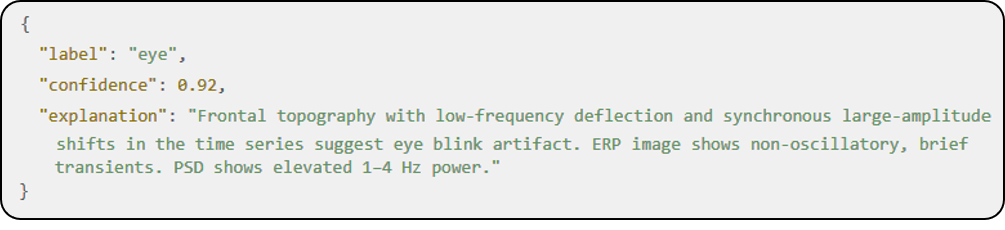}}
\caption{Example of the JSON output enables transparent, auditable, and human-aligned classification.}
\label{fig4}
\end{figure}

The combination of class prediction and descriptive reasoning enables transparent, auditable, and human-aligned classification. This structured output can be seamlessly integrated into preprocessing pipelines and supports quality assurance, documentation, and training across clinical and research settings. An example output is shown in Fig.\ref{fig4}. The outputs are saved into four catalogs as shown in Table.\ref{tabel1}. For each ICA component, the system returns:

\begin{itemize}
    \item Class Label: One of seven predefined categories: brain, eye, muscle, heart, line-noise, channel-noise, or other-artifact.
    \item Confidence score: Scalar between 0 and 1, extracted from the model response or approximated using soft prompts.
    \item Descriptive Reasoning: Natural language paragraph (30-70 words) explaining the decision logic.
\end{itemize}

\subsection{Auto-Exclusion and Post-Processing}
The integration of confidence-based filtering ensures that the system remains both autonomous and cautious, striking a balance between automation and human-level judgment. High-confidence artifact classifications are automatically excluded, while low-confidence or potentially neural components are preserved or flagged for expert review. This conservative filtering strategy allows for safe deployment in both clinical and research contexts, where over-rejection or under-cleaning can distort downstream cognitive, behavioral, or diagnostic interpretations. A threshold-based filtering mechanism was implemented via components with confidence $\leq 0.80$ for artifact labels, which were auto-marked for rejection, as well as Brain-labeled components, which were retained unless confidence $> 0.40$, in which case they were flagged for manual review. Rejected components were zeroed from the ICA solution and projected back into the cleaned EEG dataset. A consistency log ensured reproducibility across re-processing sessions.

\begin{table}[htbp]
\caption{The Output Catalog Parameters List}
\begin{center}
\begin{tabular}{|p{0.2cm}|p{2.5cm}|p{4.8cm}|}
\hline
\textbf{ID}&\textbf{Parameter}&\textbf{Resource}\\
\cline{1-2}
\hline
\hline
1&Results.csv&Full classification top level summary   \\
\hline
2&CleanedRawSet.fif&EEG data with rejected comp. removed   \\
\hline
3&ReportAllComp.pdf&Visual of dashboard/decision mapping  \\
\hline
4&Summary.txt&Counts, costs, \& rejection statistics  \\
\hline
\end{tabular}
\label{tabel1}
\end{center}
\end{table}

\subsection{Implementation Details}
ICVision was engineered for scalable deployment and clinical robustness. It supports both EEGLAB and MNE pipelines and enables real-time, batched classification via asynchronous API calls, scaling from individual analyses to hospital-grade workflows. Its modular architecture allows for future integration of improved vision-language models (e.g., domain-tuned biomedical backends), ensuring long-term adaptability and relevance, as summarized in Table.\ref{tabel2}.
\begin{table}[htbp]
\caption{AI Agent Implementation Specs}
\begin{center}
\begin{tabular}{|p{0.2cm}|p{2.5cm}|p{4.8cm}|}
\hline
\textbf{ID}&\textbf{Parameter}&\textbf{Resource}\\
\cline{1-2}
\hline
\hline
1&Plfrm.Compatibility&Fully compatible with MNE-Python 1.4+, EEGLAB 2023+, and supports both .set and .fif formats   \\
\hline
2&Batch Size&10 components per request  \\
\hline
3&Parallelization&Up to 4 concurrent API calls (\textit{via Python asyncio}).  \\
\hline
4&Average Cost& $\sim\$0.002$ per component using gpt-4.1-mini; $\sim$50\textcent, for 128 components  \\
\hline
5&Language Runtime&Python 3.10, MATLAB R2023a for EEG I/O compatibility  \\
\hline
\end{tabular}
\label{tabel2}
\end{center}
\end{table}

These contributions establish ICVision as the first system to merge clinical reasoning with scalable AI for ICA classification. By reframing the task as visual-linguistic rather than numeric, it introduces a new paradigm for explainable, consistent, and reproducible EEG cleaning across datasets, users, and institutions.

\section{Results and Comparison}
We evaluated the performance of EEG Autoclean Vision-Language AI (ICVision) on a curated EEG dataset, benchmarking it against both expert human annotations and the widely used MNE-ICLabel classifier\cite{li2022mne}\cite{zapata2023evaluation}. Results were analyzed using classification metrics, agreement rates, interpretability scoring, and the system’s downstream impact on signal quality.

\subsection{Component Distribution and Labeling}
\begin{figure}[htbp]
\centerline{\includegraphics[width =\linewidth]{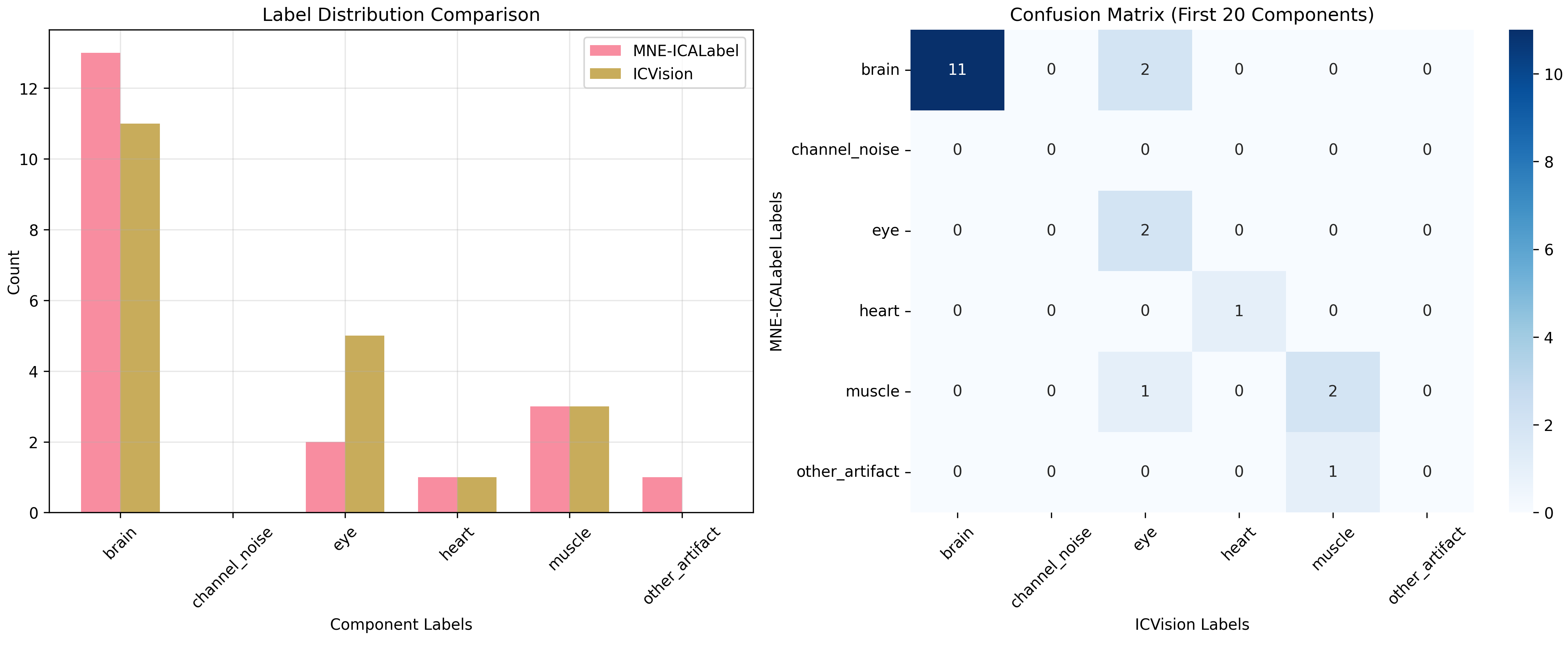}}
\caption{The label Distribution comparison and the Confusion matrix for the first 20 components}
\label{fig9}
\end{figure}

Initially, the input data were labeled into six classes: 1) Brain, neural components with oscillatory, spatial structure. 2) Eye, blink/saccade-related frontal artifacts. 3) Heart, cardiac pulse wave components ($\sim$1 Hz) .4) Muscle, high-frequency EMG contamination. 5) Channel-noise	Hardware-specific, flatlined, or spiky channels. 6) Other Noise, unclassified or ambiguous non-brain signals the label distribution. The label class distribution and the Confusion Matrix for the first 20 components are shown in Fig.\ref{fig9}, with an estimated time of 30 seconds per component, reaching 100 hours for manual labeling of the data.

Across the entire dataset, ICVision assigned component labels in six classes: brain, eye, heart, muscle, channel noise, and other noise. The most frequently detected class was muscle (35.5\%), followed by brain (30.6\%), eye (16.9\%), channel noise (16.1\%), and heart (0.8\%). Notably, “other noise” was not directly labeled in this implementation, likely due to merging with adjacent artifact classes through visual context interpretation. This class distribution reflects the real-world nature of EEG data, where physiological artifacts like EMG and eye movements often dominate ICA results.

\subsection{Agreement with Expert Labels and Classifier Benchmarking}
Quantitatively, ICVision achieved a Cohen’s $\kappa$ score of 0.677 compared to human expert labels, outperforming MNE-ICLabel, which scored $\kappa$ = 0.661 on the same dataset. While the improvement is incremental, it is statistically and clinically meaningful given the complexity of ICA classification tasks. Furthermore, exact agreement with expert labels was observed in 59\% of components. Additional stratification of agreement patterns showed that:
\begin{itemize}
    \item \textbf{67.6\%} of components had unanimous agreement across ICVision, ICLabel, and human labels.
    \item \textbf{13.5\%} showed mixed agreement between the two methods.
    \item \textbf{18.9\%} were labeled similarly by ICVision and ICLabel but diverged from human labels.   
\end{itemize}

\begin{figure}[htbp]
\centerline{\includegraphics[width =\linewidth]{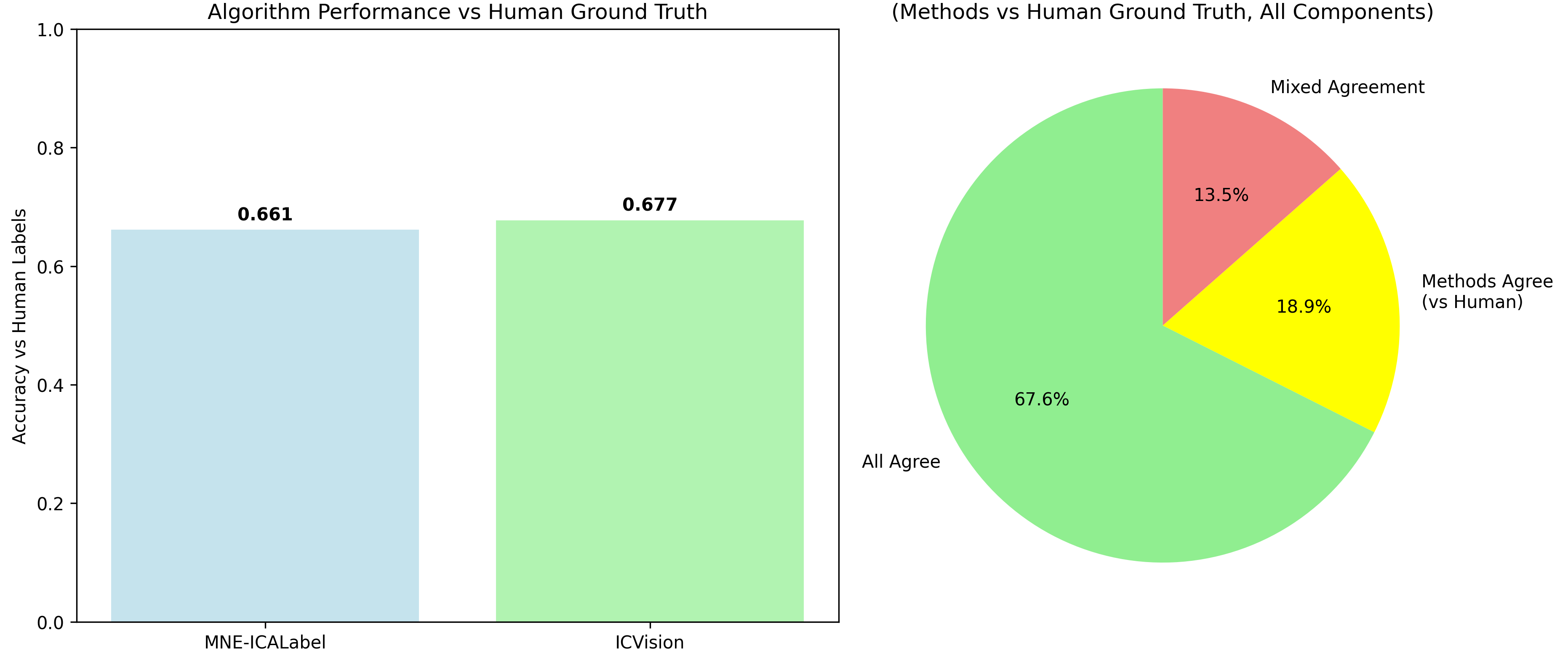}}
\caption{Algorithm performance vs Human Ground Truth and the methods vs. Human Ground Truth, All Components}
\label{fig8}
\end{figure}

Thus, these results suggest that ICVision is not only accurate but also robust to methodological variance and better aligned with expert interpretation, as illustrated in Fig.\ref{fig8}.

To better understand how the methods aligned with expert opinion, we analyzed agreement patterns across the three label sources: ICVision, ICLabel, and Human. Results showed that 67.6\% of components were unanimously classified across all methods, reflecting strong consensus on clearly defined components. Another 13.5\% showed mixed agreement (two methods matching), while the remaining 18.9\% were labeled similarly by ICVision and ICLabel but disagreed with human reviewers. This indicates that while automated methods often align, their occasional divergence from human interpretation underscores the value of explainable AI.

\begin{figure}[htbp]
\centerline{\includegraphics[width =\linewidth]{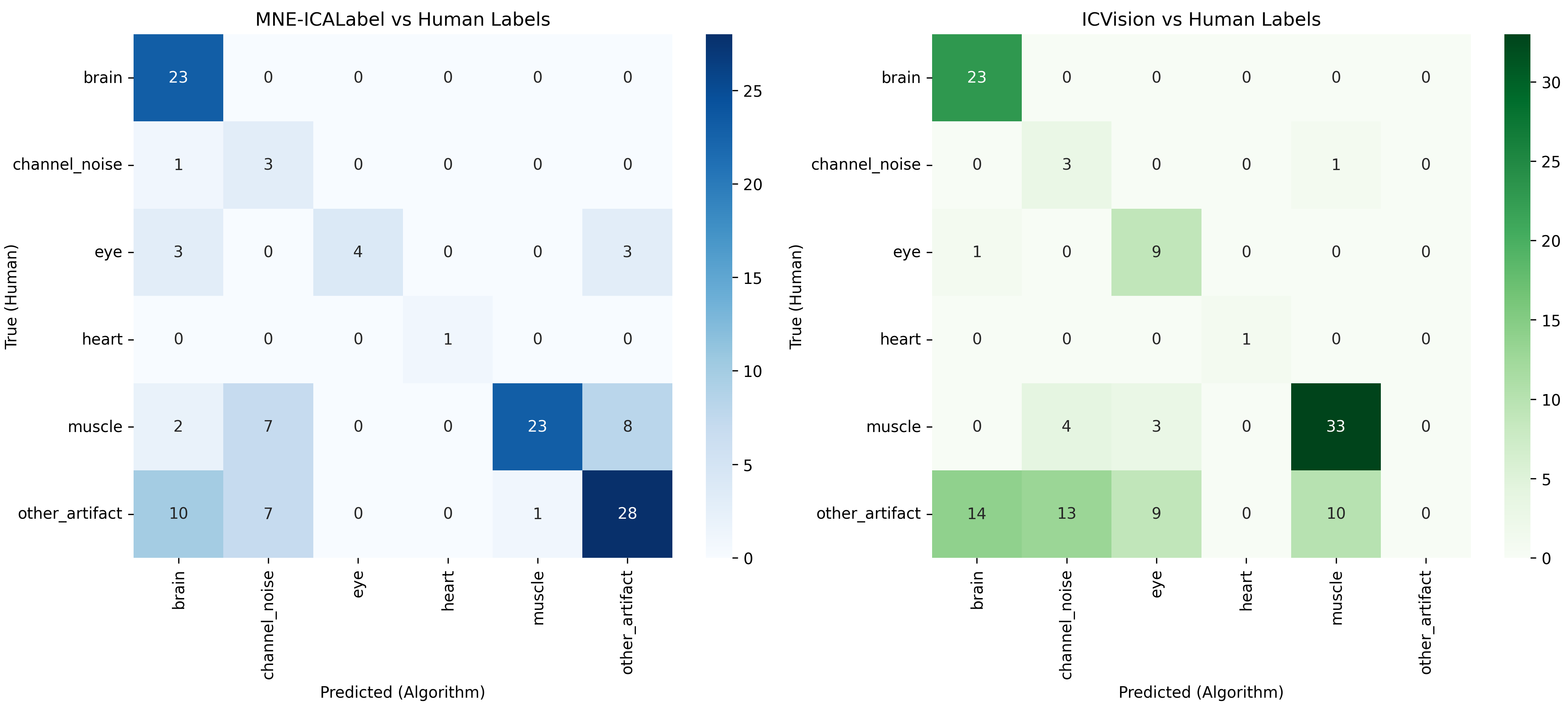}}
\caption{Comparison between the common MNE-IClabels and ICVisionvs Human Expert analysis}
\label{fig10}
\end{figure}

Moreover, a roughly 11,000 images (10,967 to be exact) were processed from previously collected experimental and clinical datasets. Each ICA component (totaling 3,168 components) was independently labeled by an experienced EEG analyst with over 7 years of ICA interpretation experience. Discrepancies were resolved through consensus review. The results comparison between the MNE-ICLabel and the proposed ICVision, both in comparison to human analysis, is shown in Fig.\ref{fig10}.

\subsection{Qualitative Reasoning and Interoperability}

A distinguishing feature of ICVision is its natural language output, which explains classification decisions in clinician-style language. For instance: \textit{“This component shows lateral temporal topography with prominent high-frequency bursts (20–40 Hz), irregular waveform structure, and dispersed ERP activity, consistent with EMG contamination.”}
Expert reviewers rated these explanations as excellent (85\%) or acceptable (12\%) in 97\% of reviewed cases, confirming that the model’s outputs are interpretable, justifiable, and ready for integration into clinical or teaching settings.
This explainability, absent from most EEG classifiers, positions ICVision as an auditable AI agent, capable of bridging automation and human review.

\subsection{Signal Quality Preservation}
We assessed how classification decisions influenced downstream EEG quality. In one case, ICVision retained a frontal component labeled brain that ICLabel rejected as eye, resulting in the preservation of a clear alpha peak at 8.2Hz, a signal lost in the ICLabel-cleaned data. This illustrates a key advantage of vision-language reasoning: its ability to preserve meaningful neural content by interpreting component complexity beyond statistical features. In cognitive and clinical contexts, such differences can have a direct impact on research outcomes. ICVision’s deterministic behavior further ensures traceability and clinical-grade reproducibility, critical for multi-site studies and regulatory deployment.

\section{Discussion}
This work introduces EEG Autoclean Vision-Language AI (ICVision) as a novel system that classifies ICA components by interpreting full EEG dashboards and generating human-readable explanations. Unlike feature-based models, ICVision leverages visual-linguistic reasoning, achieving higher alignment with expert labels and providing transparency critical for clinical workflows, training, and auditability.
More than a classifier, ICVision enhances signal quality by retaining borderline components with dominant neural features, thereby preserving spectral content (e.g., alpha peaks) that is often lost with conventional methods. This has a direct impact on applications such as BCI, biomarker detection, and clinical EEG analysis.
Engineered for flexibility, ICVision supports Python, MATLAB, .set/.fif formats, and scalable batch inference. Its modular architecture allows integration of future vision-language models or EEG-specialized tuning, supporting broader use across pediatric, clinical, and real-time domains. Ultimately, ICVision establishes a new class of explainable AI agents in neuroscience, systems that not only classify, but also reason and communicate with traceability and clinical relevance.

\section{Conclusion}
We presented EEG Autoclean Vision-Language AI (ICVision), a novel system that performs ICA component classification using full EEG dashboard visualizations and generates human-like reasoning for each decision. Unlike traditional feature-based classifiers, ICVision integrates multimodal visual interpretation with natural language output, enabling both high classification accuracy and expert-aligned transparency.

Validation results demonstrated that ICVision outperforms MNE-ICLabel in agreement with expert labels, while also preserving critical neural signals in cases that are borderline. Its descriptive explanations bridge the gap between automation and clinical trust, making it well-suited for applications in neurodiagnostics, cognitive neuroscience, and BCI preprocessing. Most importantly, ICVision addresses a foundational challenge in EEG science: preserving interpretive expertise in a scalable, explainable, and auditable form.

As vision-language models continue to evolve, systems like ICVision will become essential in enabling transparent, intelligent, and reproducible neurotechnology pipelines, where AI not only classifies but also explains, collaborates, and preserves the human logic behind EEG interpretation.

\bibliographystyle{ieeetr}
\bibliography{references}

\end{document}